%% file: main.tex
\documentclass[a4paper, twoside, 12pt]{article}

\input{template}

\begin{document}


\newcommand{\runningauthor}{Wen \textit{et al.}} 

\newcommand{\runningheadtitle}{Subtyping brain diseases}

\newcommand{\chapternumber}{16}

\newcommand{\emailaddress}{christos.davatzikos@pennmedicine.upenn.edu}

\title{Subtyping brain diseases from imaging data} 

\author[1]{Junhao Wen}
\author[2]{Erdem Varol}
\author[1]{Zhijian Yang}
\author[1]{Gyujoon Hwang}
\author[3]{Dominique Dwyer}
\author[1]{Anahita Fathi Kazerooni}
\author[4]{Paris Alexandros Lalousis} 
\author[*, 1]{Christos Davatzikos} 

\affil[1]{Center for Biomedical Image Computing and Analytics, Perelman School of Medicine, University of Pennsylvania, Philadelphia, USA}
\affil[2]{Department of Statistics, Center for Theoretical Neuroscience, Zuckerman Institute, Columbia University, New York, USA}
\affil[3]{Department of Psychiatry and Psychotherapy, Ludwig-Maximilian University, Munich, Germany}
\affil[4]{Institute for Mental Health and Centre for Human Brain Health, School of Psychology, University of Birmingham, Birmingham, United Kingdom}
\affil[*]{Corresponding author: Christos Davatzikos, \href{mailto:\emailaddress}{christos.davatzikos@pennmedicine.upenn.edu}}

\maketitle

\afterpage{\aftergroup\restoregeometry}
\pagestyle{otherpages}

\begin{abstract}
The imaging community has increasingly adopted machine learning (ML) methods to provide individualized imaging signatures related to disease diagnosis, prognosis, and response to treatment. Clinical neuroscience and cancer imaging have been two areas in which ML has offered particular promise. However, many neurologic and neuropsychiatric diseases, as well as cancer, are often heterogeneous in terms of their clinical manifestations, neuroanatomical patterns or genetic underpinnings. Therefore, in such cases, seeking a single disease signature might be ineffectual in delivering individualized precision diagnostics. The current chapter focuses on ML methods, especially semi-supervised clustering, that seek disease subtypes using imaging data. Work from Alzheimer’s Disease and its prodromal stages, psychosis, depression, autism, and brain cancer are discussed. Our goal is to provide the readers with a broad overview in terms of methodology and clinical applications. 
\end{abstract}


\section{Introduction}
\label{sec:intro}

There is growing clinical evidence that structural and functional brain development and aging take heterogeneous paths within different subsets of the human population \cite{murray_neuropathologically_2011,noh_anatomical_2014,whitwell_patterns_2007}. This heterogeneity has been relatively ignored in case-control studies analyses, yielding a limited understanding of the diversity of underlying biological processes that might give rise to similar clinical phenotypes. The advent of high throughput neuroimaging technologies and the concentrated efforts of the collection of large-scale datasets \cite{miller_multimodal_2016,petersen_alzheimers_2010} provide a unique opportunity to dissect the structural and functional heterogeneity of brain disorders in finer details and in an unbiased data-driven manner. A developing body of  work that leverages ML and neuroimaging seeks disease subtypes of neuropsychiatric and neurodegenerative disorders, including Alzheimer’s disease (AD) \cite{varol_hydra_2017,vogel_four_2021,wen_multi-scale_2021,yang_deep_2021,young_sustain_2018,zhang_bayesian_2016}, schizophrenia \cite{chand_two_2020,dwyer_brain_2018}, and late-life depression \cite{wen_multidimensional_2021}. 

Subtyping brain diseases is a clustering problem where the goal is to break down the set of patients into distinct and relatively homogeneous subgroups (i.e., subtypes). While this has been actively investigated in the computer science community, subtyping neuroimaging data is endowed with a unique set of obstacles, such as the “curse of dimensionality” and the confounding nuisance effects, such as global demographics and scanner differences. Furthermore, brain development and pathologies often progress along a continuum, e.g., from healthy state to preclinical stages to full-fledged disease \cite{yang_surreal_ICLR_2022}, thereby modeling directly in the patient domain may lead to a biased clustering solution. Thus to tackle these problems, some recent efforts  have focused on developing semi-supervised \cite{dong_chimera_2016,varol_hydra_2017,wen_multi-scale_2021,yang_deep_2021} and unsupervised clustering methods \cite{young_sustain_2018,zhang_bayesian_2016}. 
Early studies mainly focused on unsupervised clustering methods, such as K-means \cite{hamerly_learning_2004} or hierarchical clustering \cite{day_efficient_1984}, to derive data-driven subtypes using imaging data. However, such approaches directly partition the patients based on similarities/dissimilarities, potentially biased by confounding factors, such as demographics or heterogeneity caused by unrelated pathological processes. More recently, semi-supervised clustering methods \cite{dong_chimera_2016,varol_hydra_2017,wen_multi-scale_2021,yang_deep_2021} have been proposed to tackle this problem from a novel angle. To seek a pathology-oriented clustering solution, semi-supervised approaches dissect disease heterogeneity by the “1-to-{\it k}” mapping between the reference group (i.e., healthy control (CN)) and the subgroups of the patient group (i.e., the {\it k} subtypes). This approach presumably zooms into the heterogeneity of pathological processes rather than unwanted heterogeneity in general. Furthermore, confounding variations, such as demographics, are often ruled out in these approaches. 

Aiming to provide the reader in the imaging and machine learning community with a broad guideline in terms of methodology and clinical applications, we organize the remainder of this chapter as follows. In Section 2, we provide a brief overview of clustering methods, including unsupervised and semi-supervised approaches. Section 3 discusses their applications in various neurological and neuropsychiatric disorders and diseases. Section 4 concludes the paper by discussing our main observations, methodological limitations, and future directions.

\section{Methodological development using machine learning and neuroimaging } 
\label{sec:section2}
Machine learning and neuroimaging have brought unprecedented opportunities to elucidate disease heterogeneity in various brain disorders and diseases \cite{davatzikos_machine_2019}. Several trailblazing methodological papers have been recently published \cite{yang_deep_2021,young_sustain_2018,zhang_bayesian_2016}, challenging the conventional approach of patient stratification that puts all patients into the same bucket. Among these, unsupervised \cite{young_sustain_2018,zhang_bayesian_2016} and semi-supervised clustering methods \cite{yang_deep_2021} sought to derive biologically data-driven disease subtypes, but they anchor the modeling from distinct perspectives.
For conciseness, let us note that our imaging dataset contains {\it q} healthy control (CN) samples \(\boldsymbol{X}_r=[\boldsymbol{x}_1, \ldots, \boldsymbol{x}_q], \boldsymbol{X}_r \in \mathbb{R}^{p \times q}\), representing our reference group, and {\it n} patient samples (PT) \(\boldsymbol{X}_t=[\boldsymbol{x}_1, \ldots, \boldsymbol{x}_m], \boldsymbol{X}_t \in \mathbb{R}^{p \times m}\), representing the target subtype population. We denote the whole population as a matrix $\boldsymbol{X}$ that is organized by arranging each image as a vector per column \(\boldsymbol{X}=[\boldsymbol{x}_1, \ldots, \boldsymbol{x}_{q+m}], \boldsymbol{X} \in \mathbb{R}^{p \times (q+m)}\), where {\it p} is the number of features per image. We use binary labels to distinguish the patient and control groups, where 1 represents PT and -1 means CN. Disease subtyping sought to find the number of clusters ({\it k}) in the patient group that are neuroanatomically distinct while clinically relevant.

\subsection{Unsupervised clustering}
\label{sec:subsection21}
\begin{figure}[hbtp]
	\centering
		\includegraphics[width=1\textwidth]{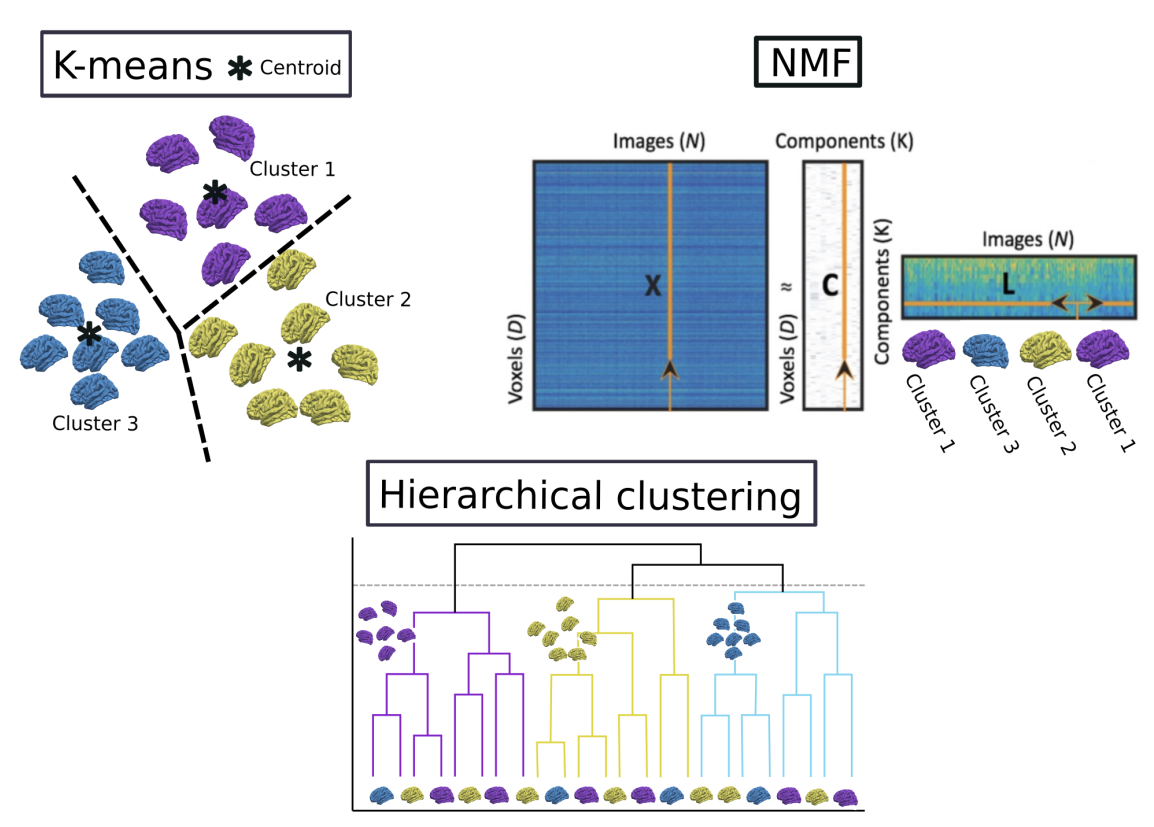}
	   \caption[Unsupervised clustering methods]{Schematic diagram of representative unsupervised clustering methods, K-means, NMF, and Hierarchical clustering. }
	\label{fig:figure1}
\end{figure}

Many recent efforts to discover the heterogeneous nature of brain diseases have investigated different unsupervised clustering algorithms \cite{ezzati_detecting_2020,honnorat_neuroanatomical_2019,jung_classifying_2016,lubeiro_identification_2016,nettiksimmons_biological_2014,Ota_stratification_2016,pan_morphological_2020,park_robust_2017,planchuelo-gomez_identificacion_2020,poulakis_fully_2020,poulakis_heterogeneous_2018,sugihara_distinct_2016,ten_kate_atrophy_2018,young_sustain_2018,zhang_bayesian_2016}. Among these approaches, the key clustering methods are often K-means, hierarchical clustering, and non-negative matrix factorization (NMF) (\autoref{fig:figure1}). In this subsection, we first briefly go through these methods. Subsequently, we focus on two representative models building on these unsupervised methods, i.e., Sustain \cite{young_sustain_2018} and latent Dirichlet allocation \cite{zhang_bayesian_2016}.
 
\subsubsection{K-means clustering}
K-means clustering aims to directly partition the {\it n} patients into {\it k} clusters. Each patient belongs to the cluster with the nearest mean (i.e., cluster centroid) quantified by a distance metric of choice (e.g., Euclidean distance). Since searching the global minimum in clustering is computationally difficult (NP-hard), local minima are searched in the K-means algorithm via an iterative refinement approach. This usually involves two steps after giving an initial set of {\it k} centroids: i) assignment step, assigning each data point to the cluster with the nearest centroid with the least squared Euclidean distance; ii) update step, recalculating means (centroids) for all data points assigned to each cluster. The two steps iteratively continue until the convergence, i.e., the assignments no longer change. More details regarding the k-means algorithm are provided in Chapter 2, Section 12.1. Please refer to \cite{feder_sample_2017,price_data-driven_2017,price_parsing_2017} for representative studies using K-means for disease subtyping.

\subsubsection{NMF clustering}
Non-negative matrix factorization (NMF) is a method that implicitly performs clustering by taking advantage that complex patterns can be construed as a sum of simple parts. In essence, the input data \(\boldsymbol{X}_{t}\) is factorized into two non-negative matrices \(\boldsymbol{C} \in \mathbb{R}^{p \times k}\) and \(\boldsymbol{L} \in \mathbb{R}^{k \times m}\), for which we refer to the component matrix and loading coefficient matrix, respectively. This method has been widely used as an effective dimensionality reduction technique in signal processing and image analysis \cite{lee_algorithms_2001}. By its nature, the \(\boldsymbol{L}\) matrix can be directly used for clustering purposes, which is analogous to K-means if we impose an orthogonality constraint on the \(\boldsymbol{L}\) matrix. Specifically, if \(\boldsymbol{L}_{kj} > \boldsymbol{L}_{ij}\) for all \(i\ne k\), this clusters the data point \(x_{n}\) into the {\it k}-th cluster. The vectors of the \(\boldsymbol{C}\) matrix indicate the cluster centroids. Please refer to \cite{ten_kate_atrophy_2018} for a representative study using NMF for disease subtyping.

\subsubsection{Hierarchical clustering}
Hierarchical clustering aims to build a hierarchy of clusters, including two types of approach: agglomerative and divisive \cite{day_efficient_1984}. In general, the merges and splits are determined greedily and presented in a dendrogram. Similarly, a measure of dissimilarity between sets of observations is required. Most commonly, this is achieved by using an appropriate metric (e.g., Euclidean distance) and a linkage criterion that specifies the dissimilarity of sets as a function of the pairwise distances of observations. Please refer to \cite{hong_multidimensional_2018,jeon_topographical_2019,nettiksimmons_biological_2014,Ota_stratification_2016,poulakis_heterogeneous_2018} for representative studies using the hierarchical clustering for disease subtyping.

\subsubsection{Representative unsupervised clustering methods}
Sustain \cite{young_sustain_2018} is an unsupervised clustering method for subtype and stage inference. Specifically, Sustain formulates the model as groups of subjects with a particular biomarker progression pattern as a subtype. The biomarker evolution of each subtype is modeled as a linear z-score model, a continuous generalization of the original event-based model \cite{young_data-driven_2014}. Each biomarker follows a piecewise linear trajectory over a common timeframe. The key advantage of this model is that it can work with purely cross-sectional data, and derive an imaging signatures of subtype and stage simultaneously.

A Bayesian latent Dirichlet allocation model \cite{zhang_bayesian_2016} was proposed to extract latent AD-related atrophy factors. This probabilistic approach hypothesizes that each patient expresses one or more latent factors, and each factor is associated with distinct but possibly overlapping atrophy patterns. However, due to the nature of latent Dirichlet allocation methods, the input images have to be discretized. Moreover, this method exclusively models brain atrophy while ignoring brain enlargement. For example, larger brain volumes in basal ganglia have been associated with one subtype of schizophrenia \cite{chand_two_2020}.

\subsection{Semi-supervised clustering}
\begin{figure}[hbtp]
	\centering
		\includegraphics[width=0.7\textwidth]{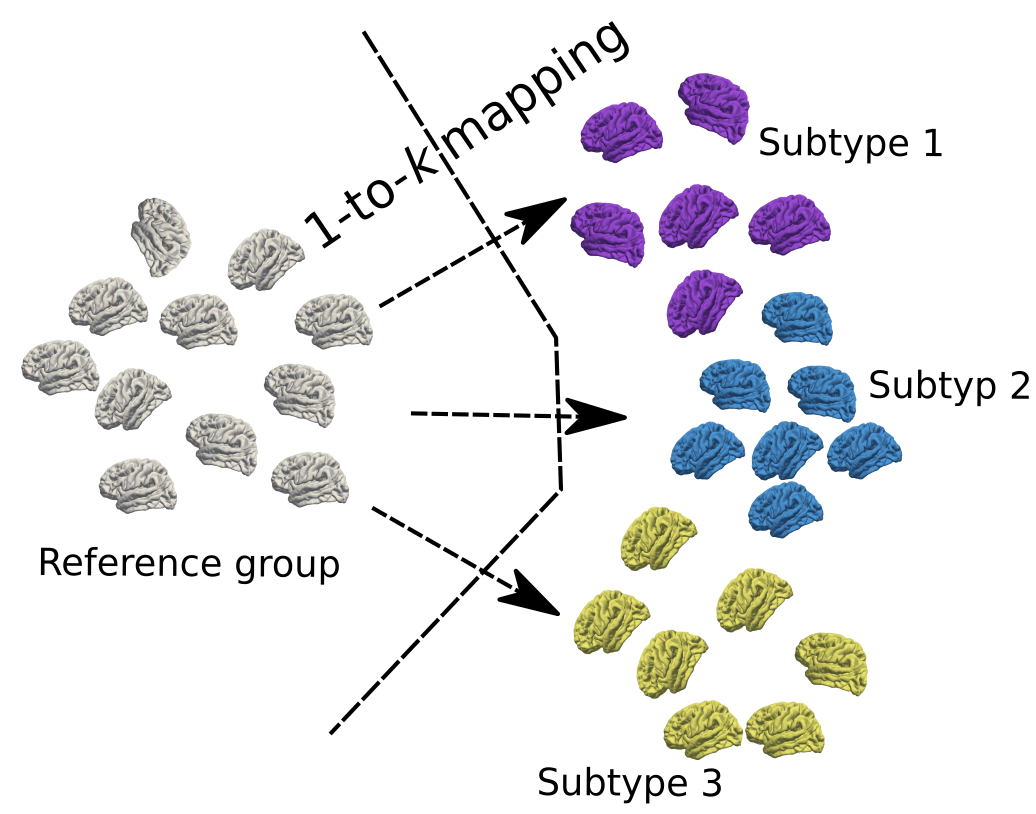}
	   \caption[semi-supervised clustering methods]{Schematic diagram of semi-supervised clustering methods. Figure is adapted from \cite{wen_multidimensional_2021}}
	\label{fig:figure2}
\end{figure}

Semi-supervised clustering methods dissect the subtle heterogeneity of interest under the principle of deriving data-driven and neurobiologically plausible subtypes (\autoref{fig:figure2}). In essence, the “1-to-{\it k}” mapping between the reference CN group and the PT group, thereby teasing out clusters that are likely driven by distinct pathological trajectories, instead of by global similarity/dissimilarity in data, which is the core momentum of conventional unsupervised clustering methods. 

In the following subsections, we briefly discuss four semi-supervised clustering methods. These methods employ different techniques to seek this “1-to-{\it k}” mapping. In particular, CHIMERA \cite{dong_chimera_2016} and Smile-GAN \cite{yang_deep_2021} utilize generative models to achieve this mapping, while HYDRA \cite{varol_hydra_2017} and MAGIC \cite{wen_multi-scale_2021} are built on top of discriminative models. 

\begin{nicebox}[Representative semi-supervised clustering methods]
    \label{box:box11}
     The central principle of semi-supervised clustering methods is to seek the "1-to-{\it k}" mapping from the reference domain to the patient domain.
    \begin{itemize}
        \item CHIMERA: a generative approach that leverages the Coherent Point Drift algorithm and maps the data distribution of the CN group to the PT group, thereby enabling to subtype by the distinct {\it k} regularized transformations.
        \item Smile-GAN: a generative approach based on GANs to learn multiple distinct mappings by generating PT from CN. Simultaneously, a clustering model is trained interactively with mapping functions to assign PT into the corresponding subtype memberships.
        \item HYDRA: a discriminative approach which leverages multiple linear support vector machines to construct a polytope that clusters the patients depending on the patterns of differences between the CN group and the PT group. 
        \item MAGIC: a generalization of HYDRA that aims to dissect disease heterogeneity at multiple imaging scales for a scale-consistent solution.
    \end{itemize}
\end{nicebox}

\subsubsection{CHIMERA}
CHIMERA employs a generative probabilistic approach, considers all samples as points in the imaging space, and infers the clusters from the transformations between the CN and PT distributions. It hypothesizes that the PT distribution can be generated from the CN distribution under {\it k} sets of transformations, each reflecting a distinct disease process. 

Mathematically, the transformation $\boldsymbol{T}$ is a convex combination of the {\it k} linear transformations that map a CN subject in the reference space to the target space: $\boldsymbol{x}_i^r \in \mathbb{R}^{q} 	\rightarrow \boldsymbol{x}_i^t =\boldsymbol{T}(x_i )=\sum_{j=1}^{k}\xi_j\boldsymbol{T}_j(x_i )$, where $\xi_j$ is the probability that a PT belongs to the {\it j}-th subtype. Ideally, if the disease subtypes were distinct, $\xi_j$ should take value 1 for the transformation corresponding to this specific disease subtype and value 0 otherwise. At its core, the Coherent Point Drift algorithm \cite{Myronenko_CPD_PAMI_2010}, a generative probabilistic approach, is used to estimate the transformation $\boldsymbol{T}$. Specifically, the CN sample point is mapped to the PT domain and regarded as a centroid of a spherical Gaussian cluster, whereas the patient points are treated as independent and identically distributed data generated by a Gaussian Mixture Model (GMM) with equal weights for each cluster. The goal is to maximize the data likelihood during the distribution matching while also taking into account covariate confounds (e.g., age and gender). The Expectation-Maximization approach is adopted to optimize the resulting energy objective. Clustering inference is straightforward after the optimized transformation \(\boldsymbol{T}_j\) is achieved, i.e., a patient can be assigned the subtype membership corresponding to the largest likelihood.

\subsubsection{Smile-GAN}
Smile-GAN is a novel generative deep learning approach based on Generative Adversarial Networks (GAN). The reader may refer to Chapter 5 for generic information about GANs. Smile-GAN aims to learn a mapping function, $f$, from joint CN domain $\mathcal{X}$ and subtype domain $\mathcal{Z}$ to the PT domain $\mathcal{Y}$, by transforming CN data $\bm{x}$ to different synthesized PT data $\bm{y'}=f(\bm{x},\bm{z})$ that are indistinguishable from real PT data, $\bm{y}$, by the discriminator, $D$. Mapping function, $f$, is regularized for inverse consistencies, with a clustering function, $g:\mathcal{Y} \rightarrow \mathcal{Z}$, trained interactively to reconstruct $\bm{z}$ from synthesized PT data $\bm{y'}$. The clustering function, $g$, can also be directly used to cluster both training and unseen test data after the training process. 

More specifically, three different data distributions are denoted as $\mathbf{x} \sim p_\text{CN}$ (for controls), $\mathbf{y} \sim p_\text{PT}$ (for patients) and $\mathbf{z} \sim P_\text{Sub}$ (for a subtype), respectively, where $\mathbf{z} \sim P_\text{Sub}$ is sampled from a discrete uniform distribution and encoded as a one-hot vector with dimension $K$ (number of clusters). Mapping function, $f:\mathcal{X}*\mathcal{Z} \rightarrow \mathcal{Y}$, and clustering function, $g:\mathcal{Y} \rightarrow \mathcal{Z}$, are learned through the following training procedure ($l_c$ denotes the cross-entropy loss): 
\begin{align}
f,g=\arg \min_{f,g} \max_{D} L_\text{GAN}(D,f)+\mu L_\text{change}(f)+\lambda L_\text{cluster}(f,g)
\end{align}
where 
\begin{align}
L_\text{GAN}(D,f)&=\mathbb{E}_{\mathbf{y}\sim p_\text{PT}}[\log(D(\bm{y}))]\notag\\&+\mathbb{E}_{\mathbf{z}\sim p_\text{Sub},\mathbf{x}\sim p_\text{CN}}[1-\log(D(f(\bm{x},\bm{z})))]]\\
L_\text{change}(f)&=\mathbb{E}_{\mathbf{x}\sim p_\text{CN},\mathbf{z}\sim p_\text{Sub}}[||f(\bm{x},\bm{z})-\bm{x}||_1]\\
L_\text{cluster}(f,g)&=\mathbb{E}_{\mathbf{x}\sim p_\text{CN},\mathbf{z}\sim p_\text{Sub}}[l_c (\bm{z},g(f(\bm{x},\bm{z})))]
\end{align}
The objective consists of adversarial loss $L_\text{GAN}$, regularization terms $L_\text{change}$ and $L_\text{cluster}$. Adversarial loss forces the synthesized PT data to follow similar distributions as real PT data. The discriminator $D$, trying to identify synthesized PT data from real PT data, attempts to maximize the loss while the mapping $f$ attempts to minimize against it. Both regularization terms serve to constrain the function class where the mapping function $f$ is sampled from so that it is truly meaningful while matching the distributions. Minimization of $L_\text{change}$ encourages sparsity of regions captured by $f$, with the assumption that only some regions are changed by disease effect. Optimizing $L_\text{cluster}$ ensures that the input sub variable $\bm{z}$ can be reconstructed from synthesized PT data $\bm{y'}$, so that the mutual information between $\bm{z}$ and $\bm{y}'$ are maximized, and distinct imaging patterns are synthesized when $\bm{z}$ takes different values. Further regularization is also imposed by forcing mapping function f and clustering function g to be Lipschitz continuous. More importantly, thanks to the inverse consistencies led by $L_\text{cluster}$, function $g$ can directly output cluster probabilities and cluster labels when given unseen test PT data. 

\subsubsection{HYDRA}
In contrast to the generative approaches used in CHIMERA and Smile-GAN, HYDRA leverages a widely used discriminative method, i.e., support vector machines (SVM), to seek this “1-to-{\it k}” mapping. The novelty is that HYDRA extends multiple linear SVMs to the non-linear case in a piecewise fashion, thereby simultaneously serving for classification and clustering. Specifically, it constructs a convex polytope by combining the hyperplane from {\it k} linear SVMs, separating the CN group and the {\it k} subpopulation of the PT group. Intuitively, each face of the convex polytope can be regarded to encode each subtype, capturing a distinct disease effect.

The convex polytope is estimated by sequentially solving each linear SVM as a sub-problem under the principle of the sample weighted SVM \cite{chang_libsvm_2011}. The optimization stops when the sample weights become stable, i.e., the polytope is stably established. The objective of maximizing the polytope’s margin can be summarized as:
\begin{align}
    \min_{\{\bm w_j, b_j\}_{(j=1)}^k}\sum_{(j=1)}^k\frac{||\bm w_j ||_2^2}{2}+\mu\sum_{i|y_i=+1}\frac{1}{k} \max\{0,1-\bm w_j^T \bm X_i^T- b_j\}\notag\\ +\mu\sum_{i|y_i=-1} s_{i,j} \max\{0,1+\bm w_j^T \bm X_i^T+ b_j\}
\end{align}
where $\bm w_j$ and $b_j$ are the weight and bias for each hyperplane, respectively. $\mu$ is a penalty parameter on the training error, and $\boldsymbol{S}$ is the subtype membership matrix of dimension $m*k$ deciding whether a patient sample {\it i} belongs to subtype {\it j}. The cluster membership is inferred as follows: 
\begin{align}
\bm S_{i,j}=\Big\{\begin{matrix}
1,j= \argmax_j(\bm w_j^T \bm X^T+ b_j)\\
0, otherwise
\end{matrix}
\end{align}

\subsubsection{MAGIC}
MAGIC was proposed to overcome one of the main limitations that HYDRA faced. That is, a single-scale set of features (e.g., atlas-based regions of interest) may not be sufficient to derive subtle differences, compared to global demographics, disease heterogeneity, since ample evidence has shown that the brain is fundamentally composed of multi-scale structural or functional entities. To this objective, MAGIC extracts multi-scale features in a coarse-to-fine granular fashion via stochastic orthogonal projective non-negative matrix factorization (opNMF) \cite{sotiras_finding_2015}, a very effective unbiased, data-driven method for extracting biologically interpretable and reproducible feature representations. Together with these multi-scale features, HYDRA is embedded into a double-cyclic optimization procedure to yield robust and scale-consistent cluster solutions. 

MAGIC encapsulates the two previous proposed methods (i.e., opNMF and HYDRA) and optimizes the clustering objective for each single-scale feature as a sub-optimization problem. To fuse the multi-scale clustering information and enforce the clusters to be scale-consistent, it adopts a double-cyclic procedure that transfers and fine-tunes the clustering polytope. First, i) inner cyclic procedure: let us remind that HYDRA decides the clusters based on the subtype membership matrix ($\bm S$). MAGIC first initializes the $\bm S$ matrix with a specific single-scale feature set, i.e., $L_i$, then the $\bm S$ matrix is transferred to the next set of feature set $L_{i+1}$ until the predefined stopping criterion is achieved (i.e., the clustering solution across scales are stable). Secondly, ii) outer cyclic procedure: the inner cyclic procedure was repeated by initializing with each single-scale feature set. Finally, to determine the final subtype assignment, we perform a consensus clustering by computing a co-occurrence matrix based on all the clustering results and then perform spectral clustering \cite{ng_spectral_2002}.

\section{Application to brain disorders}
Brain disorders and diseases affect the human brain across a wide age range. Neurodevelopmental disorders, such as autism spectrum disorders (ASD), are usually present from early childhood and affect daily functioning \cite{faras_autism_2010}. Psychotic disorders, such as schizophrenia, involve psychosis that is typically diagnosed for the first time in late adolescence or early adulthood \cite{gottesman_schizophrenia_1982}. Dementia and mild cognitive impairment (MCI)  prevail both in late mid-life for early-onset AD (usually 30 to 60 years of age) and most frequently in late-life for late-onset AD (usually over 65 years of age) \cite{mucke_alzheimers_2009}. Brain cancers in children and adults are heterogeneous and encompass over 100 different histological types of tumors, based on cells of origin and other histopathological features, and have substantial morbidity and mortality \cite{ostrom_risk_2019}. Ample clinical evidence encourages the stratification of the patients in these brain disorders and cancers, potentially paving the road towards individualized precision medicine.

This section collectively overviews previous work aiming to unravel imaging-derived heterogeneity in ASD, psychosis, major depressive disorders (MDD), MCI and AD, and brain cancer.

\subsection{Autism Spectrum Disorder}
ASD encompasses a broad spectrum of social deficits and atypical behaviors \cite{masi_overview_2017}. Heterogeneity of its clinical presentation has sparked massive research efforts to find subtypes to better delineate its diagnosis \cite{lord_annual_2012,wolfers_pattern_2019}. Recent initiatives to aggregate neuroimaging data of ASD, such as the ABIDE \cite{di_martino_enhancing_2017} and the EU-AIMS \cite{loth_eu-aims_2017}, also have motivated large-scale subtyping projects using imaging signatures \cite{hong_toward_2020}.

Different clustering methods have been applied to reveal structural brain-based subtypes, but primarily traditional techniques such as the K-means \cite{chen_parsing_2019} or hierarchical clustering \cite{hong_multidimensional_2018}. Besides structural MRI, functional MRI \cite{easson_functional_2019} and EEG \cite{duffy_autism_2019} have also been popular modalities. For reasons discussed earlier, normative clustering and dimensional analyses are better suited to parse a patient population that is highly heterogeneous \cite{tang_reconciling_2020}. However, efforts in this avenue have been primitive, with only a few recent publications using cortical thickness \cite{zabihi_fractionating_2020}. Taken together, although more validation and replication efforts are necessary to define any reliable neuroanatomical subtypes of ASD, some convergence in findings has been noted \cite{hong_toward_2020}. First, most sets of ASD neuroimaging subtypes indicate a combination of both increases and decreases in imaging features compared to the CN group, instead of pointing in a uniform direction. Second, most subtypes are characterized by spatially distributed imaging patterns instead of isolated or focal patterns. Both findings emphasize the significant heterogeneity in ASD brains and the need for better stratification. 

The search for subtypes in the ASD population has unique challenges. First, the early onset of ASD implies that it is heavily influenced by neurodevelopmental processes. Depending on the selected age range, the results may significantly differ. Second, ASD is more prevalent in males, with three to four male cases for one female case \cite{loomes_what_2017}, which adds a layer of potential bias. Third, individuals with ASD often suffer psychiatric comorbidities, such as ADHD, anxiety disorders, obsessive-compulsive disorder, among many others \cite{gillberg_autism_2014}, which, if not screened carefully, can dilute or alter the true signal. 

\subsection{Psychosis}
Psychosis is a medical syndrome characterized by unusual beliefs called delusions and sometimes hallucinations of visions, sounds, smells, or body sensations that are not present in reality. Symptoms, functioning, and outcomes are highly heterogeneous across individuals, leading to long-standing hypotheses of underlying brain subgroups. However, objective brain biomarkers have largely not been discovered for any psychosis diagnosis, stage, or clinically-defined subgroup \cite{haijma_brain_2013,pantelis_neuroanatomical_2003}. Neuroimaging studies are also affected by brain heterogeneity \cite{abi-dargham_search_2016,Insel_medicine_2015}. Recent research has thus focused on finding structural brain subtypes using unbiased statistical techniques \cite{chand_two_2020,dwyer_brain_2018,kaczkurkin_approaches_2020}.

Psychosis studies have mainly focused on determining subtypes by clustering brain structural data within the chronic schizophrenia population that has had the illness for years, with results demonstrating two \cite{chand_two_2020,dwyer_brain_2018}, three \cite{pan_morphological_2020}, and six \cite{sugihara_distinct_2016} subgroups. Various clustering techniques have been used to achieve these outcomes, including conventional approaches, such as k-means, in addition to more advanced machine learning methods, such as semi-supervised learning. A limitation of the work so far has been the lack of internal or external validation. Still, in studies with robust internal validation methods using metrics that choose the optimal cluster number based on the stability of the solution (e.g., consensus clustering), subtypes cluster along the lines of the severity of brain differences. 

In a recent study, with the largest sample to date (n=671), clustered individuals with chronic schizophrenia using HYDRA and multiple internal validation procedures were applied (i.e., cross-validation resampling, split-half reproducibility, and leave-site-out validation) \cite{chand_two_2020}. A two-subtype solution was found, with one subtype demonstrating widespread reductions and the other showing the localized larger volume of the striatum that was not associated with antipsychotic use. Interestingly, there were limited associations with current psychosis symptoms in this work, but indications of associations with education and illness duration in specific subtypes. 

Functional imaging has also been used to define psychosis subgroups using functional connectivity at rest \cite{yang_brain_2014} and effective connectivity during task performance \cite{brodersen_dissecting_2014}. The research commonly has relatively low sample sizes with little internal or external validation. Still, of these works, preliminary results demonstrate that clusters can follow diagnostic divisions between individuals with psychosis \cite{brodersen_dissecting_2014} and that specific networks (e.g., frontoparietal network) are associated with specific psychotic symptoms \cite{brodersen_dissecting_2014} \cite{yang_brain_2014}. A recent advanced deep learning approach has also revealed clinical separations along the lines of symptom severity \cite{yan_mapping_2021}. Taken together with brain structural results, it is possible that functional imaging maps onto symptom states rather than underlying illness traits that are captured by structural imaging. Further internal and external validation work is required to investigate this hypothesis by characterizing, comparing, and ultimately combining clustering solutions. A critical future direction will also be to conduct longitudinal studies that track individuals over time. Such research could lead the way towards clinical translation. 

\subsection{Major depressive disorder}
MDD is a common, severe, and recurrent disorder, with over 300 million people affected worldwide, and is characterized by low mood, apathy, and social withdrawal, with symptoms spanning multiple domains \cite{world_health_organization_icd-10_1992}. Its vast heterogeneity is exemplified by the fact that according to DSM-5 criteria, at least 227 and up to 16,400 unique symptom presentations exist \cite{fried_depression_2015,lynch_causes_2020}. The potential causes for this heterogeneity vary from divergent clinical symptom profiles to genetic aetiologies and individual differences in treatment outcomes.

Despite neurobiological findings in MDD spanning cortical thickness, gray matter volume (GMV), and fractional anisotropy (FA) measures, objective brain biomarkers that can be used to diagnose and predict disease course and outcome remain elusive \cite{buch_dissecting_2021,lynch_causes_2020,rajkowska_morphometric_1999}. Recently, there have been efforts to identify neurobiologically based subtypes of depression using a bottom-up approach, mainly using data from resting-state fMRI \cite{lynch_causes_2020}. Several studies \cite{feder_sample_2017,price_data-driven_2017,price_parsing_2017} employed k-means clustering and group iterative multiple model estimation respectively to identify two functional connectivity subtypes, while Tokuda et al. \cite{Toduka_identification_2018} and Drysdale et al. \cite{drysdale_resting-state_2017} identified three and four subtypes respectively using non-parametric Bayesian mixture models and hierarchical clustering. These subtypes are characterized by reduced connectivity in different networks, including the Default Mode Network (DMN), ventral attention network, and frontostriatal and limbic dysfunction. Regarding structural neuroimaging, one study has used K-means clustering on fractional anisotropy (FA) data to identify two depression subtypes. The first subtype was characterized by decreased FA in the right temporal lobe and the right middle frontal areas and was associated with an older age at onset. In contrast, the second subtype was characterized by increased FA in the left occipital lobe and was associated with a younger age at onset \cite{cheng_delineation_2014}.

Current research in the identification of brain subtypes in MDD has produced results that are promising but confounded by methodological and design limitations. While some studies have shown clinical promises such as predicting higher depressive symptomatology and lower sustenance of positive mood \cite{price_data-driven_2017,price_parsing_2017}, depression duration \cite{feder_sample_2017}, and TMS therapy response \cite{drysdale_resting-state_2017}, they are confounded by limitations such as relatively small samples sizes, nuisance variances such as age, gender, and common ancestry, lack of external validation, and lack of statistical significance testing of identified clusters. Furthermore, there has been a lack of ambition in the use of novel clustering techniques. Clustering based on structural neuroimaging is limited compared to other disease entities and is an avenue that future research should consider. Future studies should also aim to perform longitudinal clustering to elucidate the stability of identified brain subtypes over time and examine their utility in predicting disease outcomes.

\subsection{MCI and AD}
AD, along with its prodromal stage presenting MCI, is the most common neurodegenerative disease, affecting millions across the globe. Although a plethora of imaging studies have derived AD-related imaging signatures, most studies ignored the heterogeneity in AD. Recently, there has been a developing body of effort to derive imaging signatures of AD that are heterogeneity-aware (i.e., subtypes) \cite{vogel_four_2021,wen_multi-scale_2021,yang_deep_2021,young_sustain_2018,zhang_bayesian_2016}. 

Most previous studies leveraged unsupervised clustering methods such as Sustain \cite{young_sustain_2018}, NMF \cite{ten_kate_atrophy_2018}, latent Dirichlet allocation \cite{zhang_bayesian_2016}, and hierarchical clustering \cite{jeon_topographical_2019,nettiksimmons_biological_2014,Ota_stratification_2016,poulakis_heterogeneous_2018}. Other papers \cite{dong_heterogeneity_2016,ezzati_detecting_2020,filipovych_jointmmcc_2012,varol_hydra_2017,yang_deep_2021} utilized semi-supervised clustering methods. Due to the variabilities of the choice of databases and methodologies and the lack of ground truth in the context of clustering, the reported number of clusters and the subtypes’ neuroanatomical patterns differ and cannot be directly compared. The targeted heterogeneous population of study also varies across papers. For instance, \cite{varol_hydra_2017} focused on dissecting the neuroanatomical heterogeneity for AD patients, while \cite{dong_heterogeneity_2016} included AD plus MCI and \cite{ezzati_detecting_2020} studied MCI only. However, some common subtypes were found in different studies. First, a subtype showing a typical diffuse atrophy pattern over the entire brain was witnessed in several studies \cite{dong_heterogeneity_2016,jeon_topographical_2019,jung_classifying_2016,park_robust_2017,poulakis_heterogeneous_2018,poulakis_fully_2020,ten_kate_atrophy_2018,varol_hydra_2017,wen_multi-scale_2021,yang_deep_2021,young_sustain_2018}. Another subtype demonstrating nearly normal brain anatomy was robustly identified \cite{dong_chimera_2016,ezzati_detecting_2020,jung_classifying_2016,nettiksimmons_biological_2014,Ota_stratification_2016,poulakis_fully_2020,poulakis_heterogeneous_2018,wen_multi-scale_2021,yang_deep_2021}. Moreover, studies \cite{dong_heterogeneity_2016,poulakis_heterogeneous_2018,poulakis_fully_2020,wen_multi-scale_2021,yang_deep_2021} also reported one subtype showing atypical AD patterns (i.e., hippocampus or medial temporal lobe atrophy spared). 

Though these methods enabled a better understanding of heterogeneity in AD, there are still limitations and challenges. First, due to demographic variations and the existence of comorbidities, it is not guaranteed that models cluster the data based on variations of the pathology of interest. Semi-supervised methods might tackle this problem to some extent, but more careful sample selection and further study with longitudinal data may ensure disease-specificity. Second, spatial differences and temporal changes may simultaneously contribute to subtypes derived through clustering methods. Third, subtypes captured from neuroimaging data alone bring limited insight into  disease treatments, thereby a joint study of neuroimaging and genetic heterogeneity may provide greater clinical value \cite{wen_multidimensional_2021,chand_scz_ajp_2022}. 

\subsection{Brain cancer}
Brain tumors, such as glioblastoma (GBM), exhibit extensive inter and intra-tumor heterogeneity, diffuse infiltration, and invasiveness of various immune and stromal cell populations, which pose diagnostic and prognostic challenges, and render the standard therapies futile \cite{Sottoriva_gliblastoma_PNAS_2013}. Deciphering the underlying heterogeneity of brain tumors, which arises from genomic instability of these tumors, plays a key role in understanding and predicting the course of tumor progression and its response to the standard therapies, thereby designing effective therapies targeted at aberrant genetic alterations \cite{akhavan_car_2019,qazi_intratumoral_2017}. 
Medical imaging noninvasively portrays the phenotypic differences of brain tumors and their micro-environment caused by molecular activities of tumors on a macroscopic scale \cite{davatzikos_precision_2019,gillies_radiomics_2016}. It has the potential to provide readily accessible and surrogate biomarkers of particular genomic alterations, predict response to therapy, avoid risks of tumor biopsy or inaccurate diagnosis due to sampling errors, and ultimately develop personalized therapies to improve patient outcomes. An imaging subtype of brain tumors may provide a wealth of information about the tumor, including distinct molecular pathways \cite{fathi_kazerooni_imaging_2020,gevaert_glioblastoma_2014}. 

Recent studies on radiomic analysis of multiparametric MRI (mpMRI) scans provide evidence of distinct phenotypic presentation of brain tumors associated with specific molecular characteristics. These studies propose that quantification of tumor morphology, texture, regional microvasculature, cellular density, or microstructural properties can map to different imaging subtypes. In particular, one study \cite{rathore_radiomic_2018-1} discovered three distinct clusters of GBM subtypes through unsupervised clustering of these features, with significant differences in survival probabilities and associations with specific molecular signaling pathways. These imaging subtypes, namely solid, irregular, and rim-enhancing, were significantly linked to different clinical outcomes and molecular characteristics, including isocitrate dehydrogenase-1, O6-methylguanine–DNA methyltransferase, epidermal growth factor receptor variant III, and transcriptomic molecular subtype composition.

These studies have offered new insights into the characterization of tumor heterogeneity on both microscopic, i.e., histology and molecular, and macroscopic, i.e., imaging levels, consequently providing a more comprehensive understanding of the tumor aggressiveness and patient prognosis, and ultimately, the development of personalized treatments.  

\section{Conclusion}
Taken together, these novel clustering algorithms tailored for high resolution yet highly variable neuroimaging datasets have demonstrated a broad utility in disease subtyping across many neurological and psychiatric conditions. Simultaneously, cautions need to be taken in order not to overclaim the biological importance of subtypes, since all clustering methods find patterns in data, even if such patterns don't have a meaningful underlying biological correlate \cite{altman_clustering_2017}. External validations are necessary. For instance, evidence of post-hoc evaluations, e.g., a difference in clinical variables or genetic architectures, can support the biological relevance of identified neuroimaging-based subtypes \cite{wen_multidimensional_2021}. Moreover, good practices such as split-sample analysis, permutation tests \cite{chand_two_2020}, and comparison to the guideline of semi-simulated experiments \cite{wen_multi-scale_2021} discern the robustness of the subtypes. As dataset sizes and imaging resolution improve over time, unique computational challenges are expected to appear, along with unique opportunities to further refine our methodologies  to decipher the diversity of brain diseases. 

\section*{Acknowledgments}
This work was supported, in part, by NIH grants R01NS042645, U01AG068057, R01MH112070 and RF1AG054409. 

\bibliographystyle{spbasic}
\bibliography{references}

\end{document}

%% file: template.tex
\usepackage[noblocks]{authblk} 
\usepackage[numbers]{natbib}
\usepackage{graphicx}
\usepackage{fancyhdr}
\usepackage[colorlinks=true, linkcolor=MidnightBlue, urlcolor=MidnightBlue, citecolor=PineGreen]{hyperref}
\usepackage[tableposition=top, justification=justified, textfont=it]{caption}
\usepackage{titlesec}
\usepackage{multicol}
\usepackage{afterpage}
\usepackage{lipsum}
\usepackage[dvipsnames, x11names]{xcolor}
\usepackage[inner=4cm, outer=4cm, top=2.8cm, bottom=2.8cm, marginparsep=0cm, marginparwidth=0cm, includehead, includefoot]{geometry}
\usepackage{setspace}
\usepackage{array}
\usepackage{bm}
\usepackage{tikz}
\usepackage{float}
\usepackage[framemethod]{mdframed}
\usepackage{amsmath}
\usepackage{amssymb}

\titleformat{\section}{\Large \bfseries \centering \scshape}{\thesection.}{0.3em}{}[{\titlerule[0.5pt]}]

\definecolor{shadecolor}{RGB}{230,230,230}
\newcommand{\mybox}[1]{\par\noindent\colorbox{shadecolor}
{\parbox{\dimexpr\textwidth-2\fboxsep\relax}{#1}}}

\titleformat{\subsection}{\large \bfseries \mybox}{\thesubsection}{1em}{}

\titleformat{\subsubsection}{\itshape}{\thesubsubsection.}{0.3em}{}

\renewenvironment{abstract}
{\vskip 2.5ex {\large\bf\noindent Abstract}\vspace{0.7ex} \\ %
  \bgroup\noindent\ignorespaces}%
{\par\egroup\vskip 2.5ex}

\newenvironment{keywords}
{\bgroup\leftskip 20pt\rightskip 20pt \small\noindent{\bf Keywords:} }%
{\par\egroup\vskip 10ex}

\fancypagestyle{firstpage}{%
  
  \fancyhead[]{}
  \fancyfoot[C]{}
  \fancyfoot[L]{\footnotesize To appear in \\
  O. Colliot (Ed.), \textit{Machine Learning for Brain Disorders}, Springer\\
  }

}

\fancypagestyle{otherpages}{%
  
  \fancyhead[LE]{\thepage}
  \fancyhead[RE]{\runningauthor}
  \fancyhead[LO]{\runningheadtitle}
  \fancyhead[RO]{\thepage}
  \fancyfoot[L]{\footnotesize  \textit{Machine Learning for Brain Disorders}, Chapter~\chapternumber}
  \fancyfoot[C]{}
}

\makeatletter
\renewcommand{\maketitle}{\bgroup\setlength{\parindent}{0pt}

\begin{flushright}
  \color{MidnightBlue}
  \textbf{\LARGE Chapter~\chapternumber}
\end{flushright}

\vspace{0.3in}

\begin{flushleft}
    \setstretch{2.0} 
    \textbf{\color{MidnightBlue}\huge\@title}
\end{flushleft}

\vspace{0.15in}

\begin{flushleft}
    \textbf{\bfseries \large\@author}
\end{flushleft}\egroup
}

\makeatother

\DeclareCaptionLabelFormat{labelformat}{\color{MidnightBlue}\bfseries #1 #2}
\renewcommand{\bibpreamble}{\scriptsize \begin{multicols}{2}}
\renewcommand{\bibpostamble}{\end{multicols}}

\mdfsetup{skipabove=\topskip, skipbelow=\topskip}

\newcounter{nicebox}
\newenvironment{nicebox}[1][]{%
    \refstepcounter{nicebox}%
    \ifstrempty{#1}%
    {\mdfsetup{%
        frametitle={%
            \tikz[baseline=(current bounding box.east),outer sep=0pt]
            \node[anchor=east,rectangle,fill=blue!20]
            {\strut Theorem~\thetheo};}}
    }%
    {\mdfsetup{%
        frametitle={%
            \tikz[baseline=(current bounding box.east),outer sep=0pt]
            \node[anchor=east,rectangle,fill=blue!20]
            {\strut Box~\thenicebox:~#1};}}%
    }%
    \mdfsetup{innertopmargin=10pt,linecolor=blue!20, linewidth=2pt,topline=true, frametitleaboveskip=\dimexpr-\ht\strutbox\relax,}
    \begin{mdframed}[]\relax%
    }{\end{mdframed}}

\newfloat{floatbox}{thp}{lop}
\floatname{floatbox}{Box}

\DeclareMathOperator*{\argmax}{arg\,max}

\DeclareMathAlphabet{\mathsfit}{\encodingdefault}{\sfdefault}{m}{sl}
\SetMathAlphabet{\mathsfit}{bold}{\encodingdefault}{\sfdefault}{bx}{n}